\documentclass[letterpaper]{article} 
\usepackage{aaai23}  
\usepackage{times}  
\usepackage{helvet}  
\usepackage{courier}  
\usepackage[hyphens]{url}  
\usepackage{graphicx} 
\usepackage{natbib}  
\usepackage{caption} 
\usepackage{algorithm}
\usepackage{algorithmic}

\usepackage{newfloat}
\usepackage{listings}
\DeclareCaptionStyle{ruled}{labelfont=normalfont,labelsep=colon,strut=off} 
\floatstyle{ruled}
\newfloat{listing}{tb}{lst}{}
\floatname{listing}{Listing}
\newlength\savewidth
\newcommand{\tablestyle}[2]{\setlength{\tabcolsep}{#1}\renewcommand{\arraystretch}{#2}\centering\small}

\usepackage{booktabs}
\usepackage{multirow}
\usepackage{multicol}
\usepackage{soul}
\usepackage{url}
\usepackage{xspace}
\usepackage{xcolor}
\usepackage[colorlinks=true, citecolor=citecolor, linkcolor=linkcolor]{hyperref}
\definecolor{citecolor}{HTML}{0071BC}
\definecolor{linkcolor}{HTML}{ED1C24}

\usepackage{amsmath,amsfonts,bm}

\newcommand{\pdata}{p_{\rm{data}}}

\newcommand{\diff}{{\rm{d}}}

\def\rvfgvis{{\mathbf{f}}_{{\rm{vis}}}}
\def\rvfginvis{{\mathbf{f}}_{{\rm{inv}}}}

\def\rvbgvis{{\mathbf{b}}_{{\rm{vis}}}}
\def\rvbginvis{{\mathbf{b}}_{{\rm{inv}}}}
\def\rvmask{{\mathbf{m}}}

\def\miclub{\widehat{I}_\text{vC}}
\def\mimine{\widehat{I}_\text{M}}

\def\lossILSMI{\mathcal{L}_\text{ILS}^\text{MI}}
\def\lossILSLOne{\mathcal{L}_\text{ILS}^\text{L1}}
\def\lossILSLOneSingle{\mathcal{L}_\text{ILS}^\text{{L1-(1)}}} 
\def\lossILSLOneSeparate{\mathcal{L}_\text{ILS}^\text{L1-(2)}} 

\def\rvb{{\mathbf{b}}}

\def\rvf{{\mathbf{f}}}

\def\rvm{{\mathbf{m}}}

\def\rvx{{\mathbf{x}}}

\def\rvz{{\mathbf{z}}}

\makeatletter
\DeclareRobustCommand\onedot{\futurelet\@let@token\@onedot}
\def\@onedot{\ifx\@let@token.\else.\null\fi\xspace}
\def\eg{\emph{e.g}\onedot} 
\def\ie{\emph{i.e}\onedot} 
 
\def\etc{\emph{etc}\onedot}

\makeatother

\title{ILSGAN: Independent Layer Synthesis \\for Unsupervised Foreground-Background Segmentation}
\author {
    Qiran Zou,
    Yu Yang,
    Wing Yin Cheung,
    Chang Liu,
    Xiangyang Ji
}
\affiliations {
    Tsinghua University, BNRist\\
    qiranzou@gmail.com, yang-yu16@foxmail.com, zhangyx20@mails.tsinghua.edu.cn, \{liuchang2022, xyji\}@tsinghua.edu.cn
}

\begin{document}

\maketitle

\begin{abstract}
Unsupervised foreground-background segmentation aims at extracting salient objects from cluttered backgrounds,
where Generative Adversarial Network (GAN) approaches, especially layered GANs, show great promise.
However, without human annotations, they are typically prone to produce foreground and background layers with non-negligible semantic and visual confusion, dubbed ``information leakage", resulting in notable degeneration of the generated segmentation mask.
To alleviate this issue, we propose a simple-yet-effective explicit layer independence modeling approach, termed Independent Layer Synthesis GAN (ILSGAN), pursuing independent foreground-background layer generation by encouraging their discrepancy.
Specifically, it targets minimizing the mutual information between visible and invisible regions of the foreground and background to spur interlayer independence.
Through in-depth theoretical and experimental analyses, we justify that explicit layer independence modeling is critical to suppressing information leakage and contributes to impressive segmentation performance gains. 
Also, our ILSGAN achieves strong state-of-the-art generation quality and segmentation performance on complex real-world data.
Our code is available at: \url{https://github.com/qrzou/ILSGAN}
\end{abstract}

\section{Introduction}
Foreground-background segmentation, also called foreground extraction, targets dividing foregrounds containing object instances from the background in an image. It is an elementary and special case of instance segmentation~\cite{he2017mask_Mask_RCNN} that the ability to distinguish foreground-background is vital for complex scene understanding~\cite{krishna2017visual_VisualGenome,zellers2018neural_NeuralMotifs,yang2018graph_GraphRCNN}. Since it is error-prone, time-consuming, and expensive to densely label a large quantity of data, unsupervised foreground-background segmentation, including GAN-based~\cite{melas2021finding_FUIS,yang2017lr_LR-GAN,bielski2019emergence_PerturbGAN}, VAE-based~\cite{yu2021unsupervised_DRC,engelcke2021genesis_GENESISv2}, and MI clustering-based~\cite{ji2019invariant_IIC,ouali2020autoregressive_AR-ACL} methods, attracts growing attention in the community.

\begin{figure}[!t]
   \centering
   \includegraphics[width=\linewidth]{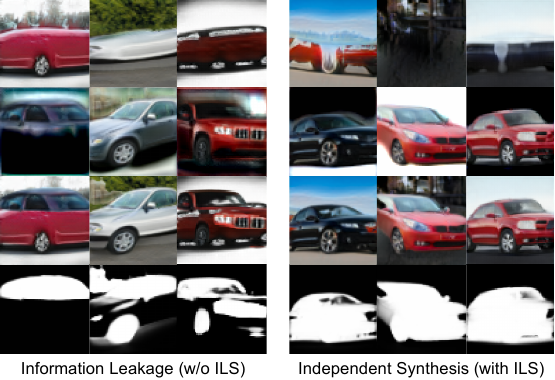}
   \caption{From top to bottom: synthetic background, foreground, composed image, and corresponding mask. 
   Left: synthetic samples with information leakage problem when our ILS algorithm is not adopted. Right: generated samples with independent foreground and background layers and more accurate masks when ILS is employed.
   }
   \label{fig:intro_fig}
\end{figure}

Among these approaches, GAN-based methods show significant representation disentanglement ability which contributes to accurate segmentation performance.
Leveraging the intrinsic compositional nature of natural images, layered GANs have made exceptional progress for unsupervised foreground-background segmentation.
Specifically, with multilayer structures of generators and corresponding inductive constraints, PerturbGAN~\cite{bielski2019emergence_PerturbGAN} considers that the foreground can be shifted to a certain extent without affecting the image fidelity.
FineGAN~\cite{singh2019finegan_FineGAN} constrains the background layer with an extra background discriminator.

However, during layered GANs' training, foreground and background layers typically contain increasing semantic information about each other, resulting in their growing visual similarity.
As shown in Fig.~\ref{fig:intro_fig} (left), there exists semantic confusion that background contains a wheel and foreground contains trees, and visual similarity that foreground and background have similar colors and textures.
This phenomenon, termed ``information leakage", discourages the mask layer from distinguishing the foreground and background and impairs segmentation performance severely. Though, it is still largely under-studied by now.

Intuitively, the semantic and visual confusion can be characterized as the excessive correlation between foreground and background layers, indicating a lack of independence.
Therefore, a feasible way to solve information leakage problem is to enhance the independence between layers.
However, such independence has not been valued in previous works, which restricts the ability of layered GANs to distinguish foreground from background.

In this work, we propose \textbf{I}ndependent \textbf{L}ayer \textbf{S}ynthesis \textbf{GAN} (ILSGAN), based on layered GAN and fueled by information theory, which generates images composed of independent layers and corresponding high-quality masks for unsupervised foreground-background segmentation.
Specifically, ILSGAN minimizes the mutual information (MI) between foreground and background layers to improve the interlayer independence.
To circumvent the difficulty of deriving the sample density in MI, it employs a variational upper bound and reduce MI by minimizing the upper bound.
Furthermore, ILSGAN can specifically optimize the mask layer to learn partitions with less MI, which effectively enhances segmentation performance.

We perform experiments on various datasets including Cars~\cite{krause20133d_Cars}, CUB~\cite{WahCUB_200_2011_CUB}, and Dogs~\cite{khosla2011novel_Dogs}.
Our ILSGAN successfully suppresses information leakage and efficiently generates complete and accurate segmentation masks.
Compared with existing unsupervised segmentation methods, ILSGAN shows a significant advantage in segmentation performance. 

We summarize our contributions as follows:
\begin{itemize}

    \item We find that interlayer semantic and visual confusion, termed information leakage, exists in layered GAN and undermines the segmentation performance.

    \item We propose ILSGAN, improving interlayer independence by minimizing mutual information, which suppresses information leakage while powerfully enhancing segmentation performance.

    \item We achieve solid state-of-the-art performances on commonly used unsupervised foreground-background segmentation datasets, with IoU results of 81.4, 72.1, and 86.3 on Cars, CUB and Dogs, beating the closet competitors (70.8, 69.7, 72.3) with significant margins.

\end{itemize}

\section{Related Work}

\subsection{Generative Adversarial Networks}
GANs~\cite{goodfellow2014generative_GAN} provide an unsupervised way for learning the representation of data and generating high-fidelity synthetic images. They have been successfully used for image editing~\cite{ling2021editgan_EditGAN,kim2021exploiting_StyleMapGAN}, style transfer~\cite{karras2019style_StyleGAN1,karras2020analyzing_StyleGAN2}, controllable image generation~\cite{chen2016infogan_InfoGAN,singh2019finegan_FineGAN,benny2020onegan_OneGAN}, representation learning~\cite{donahue2019large_AdvRepLearning}, image classification~\cite{chen2016infogan_InfoGAN}, super-resolution~\cite{ledig2017photo_SRGAN}, and generating annotations~\cite{zhang2021datasetgan_DatasetGAN,yang2021learning_AnnoPart,abdal2021labels4free_labels4free}.

Unlike traditional GANs, a series of works~\cite{chen2016infogan_InfoGAN,singh2019finegan_FineGAN,benny2020onegan_OneGAN,yang2022learning_LFBS} disentangle GANs by decomposing the input latent into a main subset and a controlling subset for disentanglement. 
The controlling subset can be used to classify the synthetic images and disentangle the semantic meaning like rotation and width.

On the other hand, layered GANs disentangle the representation explicitly by introducing a multi-layer structure that corresponds to the elements in nature scenes. They first generate multiple layers and then blend these layers into a composed image. 
It gives layered GANs the ability to decompose visual scenes and makes it possible for layered GANs to achieve fine-grained controllable generation~\cite{singh2019finegan_FineGAN,benny2020onegan_OneGAN,yang2022learning_LFBS}, video generation~\cite{ehrhardt2020relate_RELATE}, and unsupervised segmentation~\cite{yang2017lr_LR-GAN,bielski2019emergence_PerturbGAN}.

\subsection{Unsupervised Foreground-Background Segmentation}
A typical line of recent unsupervised methods partition images into foreground and background with layered GANs.
Some works spur foreground extraction with foreground perturbation~\cite{yang2017lr_LR-GAN,bielski2019emergence_PerturbGAN,yang2022learning_LFBS} or shape prior~\cite{kim2021unsupervised_ShapePrior}, distinguish foreground and background with an additional foreground discriminator~\cite{singh2019finegan_FineGAN,benny2020onegan_OneGAN}, or decompose images by redrawing real images~\cite{chen2019unsupervised_ReDO}.
These methods show impressive gains in segmentation performance. However, the independence between foreground and background is overlooked which leads to information leakage and degradation of segmentation performance.

On another front, methods based on clustering attain unsupervised segmentation through self-supervised learning with mutual information maximization~\cite{ji2019invariant_IIC,ouali2020autoregressive_AR-ACL} or contrastive learning~\cite{cho2021picie_PiCIE,van2021unsupervised_COMP,choudhury2021unsupervised_PartDiscovery,hwang2019segsort_SegSort}.
However, when these methods are applied to binary segmentation of foreground and background, they show disadvantages~\cite{benny2020onegan_OneGAN} in segmentation performance compared to methods based on layered GANs.

\begin{figure*}[t]
    \centering
    \includegraphics[width=\linewidth]{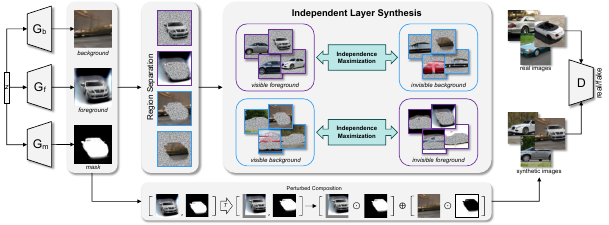}
    \caption{Framework of ILSGAN. $G_b$, $G_f$ and $G_m$ are the generators sharing latent code $z$ and producing background, foreground, and mask layers. The generated layers are fed into two branches. One branch is the perturbed composition and adversarial learning against a discriminator $D$. The other includes separating regions according to visibility in the final image and our Independent Layer Synthesis algorithm which maximizes the independence between visible and invisible regions of foreground and background layers.
    }
    \label{fig:framework}
 \end{figure*}

Apart from these works, other works extract foreground by modeling with VAE~\cite{yu2021unsupervised_DRC,burgess2019monet_MONET,engelcke2019genesis_GENESISv1,engelcke2021genesis_GENESISv2,locatello2020object_SlotAtt}, searching directions in latent space~\cite{melas2021finding_FUIS,voynov2021object_OSWL}, or maximizing inpainting error~\cite{savarese2021information_IEM}. 
Our methods leverage the high-fidelity and foreground extraction ability of layered GANs and at the same time improve interlayer independence which contributes to more accurate and clear segmentations.

\section{Method}
\subsection{Layer Synthesis}

Our ILSGAN synthesizes independent foreground, background, and mask layers with a layered structure, which are finally blended into photo-realistic images and corresponding binary masks.
Formally, let $G_f$, $G_b$, and $G_m$ denote a foreground generator, a background generator, and a mask generator, respectively.
Given a latent variable $\rvz \in \mathcal{Z}$ drawn from a prior distribution, $\rvz\sim{p_{\rm{z}}}$, the generation process can be written as
\begin{equation}\label{eq:composition}
\renewcommand{\arraystretch}{1.3}
\begin{array}{c}
    \rvf = G_f(\rvz),~~\rvb = G_b(\rvz),~~\rvmask=G_m(\rvz),\\
    \rvx = \rvmask \odot \rvf + (\bm{1} - \rvmask) \odot \rvb,
\end{array}
\end{equation}
where $\rvf$, $\rvb$, and $\rvmask$ denote foreground, background, and mask, respectively, which constitutes an image $\rvx$ by a composition step (Eq.~\ref{eq:composition}).

The full synthesis process is wrapped as $G:\mathcal{Z}\rightarrow\mathcal{X}$, which can be learned via adversarial learning against a discriminator $D:\mathcal{X}\rightarrow[0,1]$ to tell real or fake images as 
\begin{equation}\label{eq:adversarial}
\begin{aligned}
\min_G \max_D~~V(G, D) =~
& \mathbb{E}_{\rvx\sim{\pdata}}\left[\log{D(\rvx)}\right] + \\ 
& \mathbb{E}_{\rvz\sim{p_{\rm{z}}}}\left[\log{(1-D(G(\rvz)))}\right],
\end{aligned}
\end{equation}
where $\mathcal{Z}\subset\mathbb{R}^{d}$ and $\mathcal{X}\subset\mathbb{R}^{3\times H\times W}$ denote the latent variable space and the image space.

Since vanilla adversarial learning (Eq.~\ref{eq:adversarial}) lacks incentive for non-trivial layerwise generation, inductive biases such as network structure, hand-crafted regularization, \etc are essential. Hence, we briefly review the inductive biases inherited from prior work as follows. 
First, the generators are parameterized with deep convolutional neural networks. Concretely, $G_f$ and $G_b$ are different networks; $G_m$ shares backbone networks with $G_f$ and has a separate head to output masks. More details about networks structure are available in Sec.~\ref{sec:exp_detail}.
Second, to counter that $G_m$ constantly outputs $\rvmask=\bm{1}$, we replace composition step in Eq.~\ref{eq:composition} with perturbed composition~\cite{bielski2019emergence_PerturbGAN},
\begin{equation}
    \rvx = T(\rvmask) \odot T(\rvf) + 
           (1 - T(\rvmask)) \odot \rvb,
\end{equation}
where $T$ is an operation that randomly shifts masks and foreground pixels with a small amount. Third, $G_m$ is penalized if it outputs masks that a have smaller area than a pre-defined threshold. Finally, binary masks are encouraged with an additional loss function. The loss functions are defined as follows, 
\begin{equation}
\begin{aligned}
    & \mathcal{L}_{\text{m}} = 
    \mathbb{E}_{\rvz\sim{p_{\rm{z}}}, \rvmask=G_m(\rvz)}~
    \max\{0, \eta - \frac{1}{HW} \lVert \rvmask \rVert_1\}, \\
    & \mathcal{L}_{\text{b}} = 
    \mathbb{E}_{\rvz\sim{p_{\rm{z}}}, \rvmask=G_m(\rvz)}~
    \frac{1}{HW} \sum_{u\in\Omega} \min\{m_u, 1 - m_u\},
\end{aligned}
\end{equation}
where $\mathcal{L}_{\text{m}}$ and $\mathcal{L}_{\text{b}}$ respectively denote the aforementioned loss functions that encourage non-zero and binary masks. $\eta$ represents the tolerable minimal mask area, $m_u$ denotes the element in $\rvmask$ indexed by $u$ and $\Omega=\{1, \dots, H\}\times\{1, \dots, W\}$ denotes the index set.

\subsection{Independent Layer Synthesis}
We propose independent layer synthesis (ILS) loss, $\mathcal{L}_{\text{ILS}}$, to improve the independence between foreground and background layers and thereby suppress information leakage. 
ILS loss, as well as aforementioned losses, are integrated into GAN training, resulting in our ILSGAN framework which is illustrated in Fig.~\ref{fig:framework}.
The optimization of generator and discriminator are alternated~\cite{goodfellow2014generative_GAN} as follows,
\begin{equation}
\begin{array}{ll}
\min\limits_D & -V(\overline{G}, D) + \gamma R_1(D), \\
\min\limits_G & V(G, \overline{D}) + 
                \lambda_{\text{m}}\mathcal{L}_{\text{m}} + 
                \lambda_{\text{b}}\mathcal{L}_{\text{b}} + 
                \lambda_{\text{ils}}\mathcal{L}_{\text{ILS}},
\end{array}
\end{equation}
where $V(\overline{G}, D)$ and $V(G, \overline{D})$ respectively denote the variants of $V(G, D)$ when $G$ or $D$ is fixed, $R_1(D)$ denotes the gradient penalty~\cite{mescheder2018training} that stabilizes the GAN training with weight $\gamma$. And $\lambda_{\text{m}}$, $\lambda_{\text{b}}$, and $\lambda_{\text{ils}}$ denote the corresponding loss weights.  
The construction of ILS loss is described below.

\paragraph{Measuring layer independence with MI}
As observed in practice, information leakage issue (Fig.~\ref{fig:intro_fig}) frequently occurs in the unsupervised learning of layer synthesis.
This problem can be formally stated as the unexpected low interlayer independence.
Therefore, to suppress information leakage, we are motivated to increase the independence between foreground and background layers.
By information theory, this independence can be measured with mutual information (MI),
\begin{equation}\label{eq:mutual_info}
\begin{aligned}
      I(\rvb; \rvf) 
    = & \int p(\rvb, \rvf) \log{ \frac{p(\rvb, \rvf) }{ p(\rvb)p(\rvf)} } ~ \diff\rvb\diff\rvf \\
    = & H(\rvb) + \mathbb{E}_{p(\rvb, \rvf)} \log{ p(\rvb | \rvf) },
\end{aligned}
\end{equation}
where $\rvb,\rvf$ are drawn from the joint probability $p(\rvb, \rvf)$ induced by our layer synthesis model. 
To this end, one straightforward way to increase interlayer independence is to directly minimize $I(\rvb; \rvf)$.
Nonetheless, estimation of $I(\rvb; \rvf)$ is challenging since the implicit model (\ie GAN) on which our layer synthesis model is built only supports the sampling process and is unable to yield density estimation of samples.
Therefore, we instead appeal to minimizing the upper bound of $I(\rvb; \rvf)$ as follows.

\paragraph{Minimizing MI upper bound}
To circumvent the difficulty of estimating sample density, \ie $p(\rvb, \rvf)$, $p(\rvb)$, and $p(\rvf)$ in Eq.~\ref{eq:mutual_info}, 
we consider the variational CLUB~\cite{cheng2020club_CLUB} upper bound of MI,
\begin{equation}\label{eq:club_definition}
\begin{aligned}
    I(\rvb; \rvf) \leq I_{\text{vC}}(\rvb,\rvf)  
    = & \mathbb{E}_{p(\rvb, \rvf)} \left[\log{ q(\rvb | \rvf)}\right] \\
    & - \mathbb{E}_{p(\rvb)}\mathbb{E}_{p(\rvf)} \left[\log{ q(\rvb | \rvf)}\right],
\end{aligned}
\end{equation}
where $q(\rvb|\rvf)$ is a variational approximation of conditional probability $p(\rvb | \rvf)$.
This inequality Eq.~\ref{eq:club_definition} holds under mild conditions~\cite{cheng2020club_CLUB}. 
It is noteworthy that computing $I_{\text{vC}}(\rvb,\rvf)$ only requires samples from generative models and estimating density under the variational distribution, significantly easing the computation.
We further choose a Laplace distribution with identity covariance~\cite{savarese2021information_IEM} as the variational distribution, \ie $q(\rvb|\rvf) = \mathcal{L}(\rvb;\rvf, \mathbf{I}) \propto \exp(-\| \rvb - \rvf \|_1)$. 
We have 
\begin{equation}
\begin{aligned}
    I_{\text{vC}}(\rvb,\rvf) = & \mathbb{E}_{p(\rvb, \rvf)} \left[- \| \rvb - \rvf \|_1\right] \\
      & - \mathbb{E}_{p(\rvb)}\mathbb{E}_{p(\rvf)} \left[- \| \rvb - \rvf \|_1\right] + \text{Const}.
\end{aligned}
\end{equation}
Given a mini-batch of samples $\{(\rvb^{(i)},\rvf^{(i)})\}_{i=1}^N$, neglecting the constant term, the unbiased $N$-sample estimation of $I_{\text{vC}}(\rvb,\rvf)$ writes
\begin{equation}
    \miclub(\rvb,\rvf) = \frac{1}{N}\sum_{i=1}^{N} - \| \rvb^{(i)} - \rvf^{(i)} \|_1 + \| \rvb^{(k_i)} - \rvf^{(i)} \|_1,
\end{equation}
where $N$ denotes the number of samples, $\rvb^{(i)}=G_b(\rvz^{(i)})$ and $\rvf^{(i)} = G_f(\rvz^{(i)})$ are generated layers by forwarding randomly sampled latent variables $\rvz^{(i)}\sim p_z$ to our generative models, and $k_i$ denotes a uniformly sampled index in $\{1,2,...,N\}$. 
To this end, we minimize $\miclub(\rvb,\rvf)$ to indirectly reduce the $I(\rvb; \rvf)$.

\paragraph{ILS Loss} 
Moreover, in order to optimize the mask to find the segmentation with less mutual information, we separate foreground and background layers respectively according to their visibility in synthetic image $\rvx$, Fig.~\ref{fig:framework},
\begin{equation}
\renewcommand{\arraystretch}{1.3}
\begin{array}{lll}
    &\rvfgvis = \rvf \odot \rvmask,&\rvfginvis = \rvf \odot (\bm{1} - \rvmask),\\
    &\rvbgvis = \rvb \odot (\bm{1} - \rvmask),&\rvbginvis = \rvb \odot \rvmask,
\end{array}
\end{equation}
where $\rvfgvis$, $\rvfginvis$, $\rvbgvis$, and $\rvbginvis$ denote the visible and invisible regions of foreground and background layers. 
Thus, our ILS loss minimizes MI of two pairs of visible and invisible regions,
\begin{equation}\label{eq:ils_inv_vis}
    \mathcal{L}_\text{ILS} = I(\rvbginvis;\rvfgvis) + I(\rvbgvis;\rvfginvis).
\end{equation}
Estimating MI with $\miclub$, our ILS-MI loss is as follows,
\begin{equation}\label{eq:ILSLossMI}
\renewcommand{\arraystretch}{2.0}
\begin{array}{lll}
    \lossILSMI = \miclub(\rvbginvis;\rvfgvis) + \miclub(\rvbgvis;\rvfginvis) \\
     = \frac{1}{N}\sum_{i=1}^{N} \left[ - \| \rvbginvis^{(i)} - \rvfgvis^{(i)} \|_1 + \| \rvbginvis^{(k_i)} - \rvfgvis^{(i)} \|_1 \right] \\
     + \frac{1}{N}\sum_{j=1}^{N} \left[ - \| \rvbgvis^{(j)} - \rvfginvis^{(j)} \|_1 + \| \rvbgvis^{(k_j)} - \rvfginvis^{(j)} \|_1 \right].
\end{array}
\end{equation}

\subsection{Discussion}

As the information leakage problem can be interpreted as there existing unexpected interlayer visual similarity, \eg car body textures or wheels present in the background layer (Fig.~\ref{fig:intro_fig}), 
one intuitive way to suppress this issue is to directly minimize the visual similarity between foreground and background layers.
Based on this motivation, by measuring the visual similarity using $L1$ distance, the ILS-L1 loss can be constructed,
\begin{equation}
    \lossILSLOne = - \| \rvbginvis - \rvfgvis \|_1 
     - \| \rvbgvis - \rvfginvis \|_1.
\end{equation}

In practice, we observe that ILS-L1 can help to mitigate the information leakage issue and improve the segmentation performance. 
Moreover, ILS-L1 is also able to contribute a bit to the decrease of MI between foreground and background layers.
Despite so, we believe the effect of this intuitive method is limited and our ILS-MI loss is a more fundamental approach that achieves more impressive results.
See Sec.~\ref{sec:exp_analysis} for detailed experiments.

\section{Experiments}

\begin{table*}[t]\centering
	\tablestyle{6pt}{1.25}\begin{tabular}{lc ccc ccc ccc}
		\toprule
		\multirow{2}[2]{*}{Methods} & \multirow{2}[2]{*}{Sup.} & \multicolumn{3}{c}{Cars} & \multicolumn{3}{c}{CUB} & \multicolumn{3}{c}{Dogs} \\
		\cmidrule(lr){3-5} \cmidrule(lr){6-8} \cmidrule(lr){9-11}
		& & ~FID$\downarrow$ & IoU$\uparrow$ & DICE$\uparrow$ 
		& ~FID$\downarrow$ & IoU$\uparrow$ & DICE$\uparrow$ 
		& ~FID$\downarrow$ & IoU$\uparrow$ & DICE$\uparrow$ \\
		\cmidrule(lr){1-11}
		FineGAN~\cite{singh2019finegan_FineGAN}$^\dagger$  & Weak & 24.8 & 53.2 & 60.3 & 23.0 & 44.5 & 56.9 & 54.9 & 48.7 & 59.3 \\
		OneGAN~\cite{benny2020onegan_OneGAN} & Weak & 24.2 & 71.2 & 82.6 & 20.5 & 55.5 & 69.2 & \textbf{48.7} & 71.0 & 81.7 \\
		IEM~\cite{savarese2021information_IEM} & Uns. & - & - & - & - & 55.1 & 68.7 & - & - & - \\
		ReDO~\cite{chen2019unsupervised_ReDO}$^\ddagger$ & Uns. & - & 52.5 & 68.6 & - & 47.1 & 61.7 & - & 52.8 & 67.9 \\
		Impr.LGAN~\cite{yang2022learning_LFBS} & Uns. & 19.0 & 64.6 & - & 12.9 & 69.7 & - & 59.3 & 61.3 & - \\
		DRC~\cite{yu2021unsupervised_DRC} & Uns. & - & 70.8 & 82.5 & - & 54.6 & 69.4 & - & 72.3 & 83.6 \\
		\cmidrule(lr){1-11}
		ILSGAN (Ours) & Uns. & \textbf{9.0} & \textbf{81.2} & \textbf{89.3} & \textbf{8.7} & \textbf{72.1} & \textbf{82.8} & 49.5 & \textbf{86.3} & \textbf{92.5} \\
		\bottomrule
	\end{tabular}
    \caption{Quantitative comparison to related methods with respect to generation quality and segmentation. 
    $^\dagger$: Results are reported by OneGAN~\cite{benny2020onegan_OneGAN}. 
    $^\ddagger$: Results are reported by DRC\cite{yu2021unsupervised_DRC}. }
    \label{tab:compare_others}
\end{table*}

\begin{figure*}[t]
    \centering
    \includegraphics[width=\linewidth]{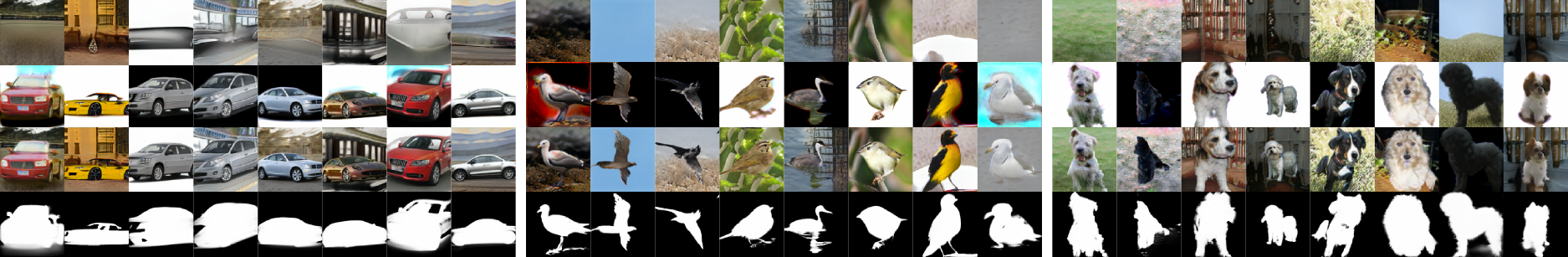}
    \caption{Qualitative generation results (128$\times$128). Left: Cars, middle: CUB, right: Dogs. From top to bottom: background, foreground, composed image, and foreground mask.}
    \label{fig:generation_results}
 \end{figure*}
 
 \begin{figure*}[t]
    \centering
    \includegraphics[width=\linewidth]{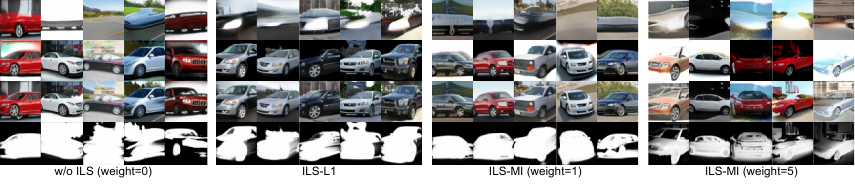}
    \caption{Qualitative ablation results of ILSGAN for settings in Table.~\ref{tab:weight_and_L1_ablation} on Cars. From top to bottom: background, foreground, composed image, and corresponding mask.}
    \label{fig:ablations_fig}
 \end{figure*}
 
\subsection{Settings}
\subsubsection{Datasets}
We evaluate our method on Stanford Cars (Cars)~\cite{krause20133d_Cars}, Stanford Dogs (Dogs)~\cite{khosla2011novel_Dogs}, and Caltech-UCSD Birds 200-2011 (CUB)~\cite{WahCUB_200_2011_CUB}.
We follow the train-test split from~\cite{yu2021unsupervised_DRC} to split Cars and Dogs dataset, leading to 6218 training images and 6104 testing images for Cars as well as 3286 training and 1738 testing images for Dogs.
The CUB dataset is split into 10k images for training and 1k images for testing following~\cite{chen2019unsupervised_ReDO}.
We use the approximated ground-truth by~\cite{yu2021unsupervised_DRC} on Cars and Dogs, and manually labeled ground-truth masks on CUB to evaluate the performance on test set.

\subsubsection{Evaluation}
As our ILSGAN is only able to generate images with segmentation masks, we train a post hoc UNet~\cite{ronneberger2015u_U-Net} on these synthetic data at 64$\times$64 resolution to evaluate its segmentation performance.  
For segmentation evaluation at 128$\times$128 resolution, we upsample the predicted masks to 128$\times$128 for computing metrics.
The trained UNet is tested on real test set and the DICE coefficient and intersection over union (IoU) metrics are reported.
To more accurately measure the independence between synthesized foreground and background layers during test phase, we employ a heavier neural estimator, MINE~\cite{belghazi2018mine_MINE}, which is a lower bound of mutual information and can serve as a reasonable MI estimator. 
Similar to Eq.~\ref{eq:ils_inv_vis}, we respectively compute the MI of regions divided by segmentation masks and sum them up, which is reported as ``MI'' in the following sections.  
Details are available in the appendix.
Besides, we also compute the Fréchet Inception Distance (FID)~\cite{heusel2017gans_FID} of 10k synthetic images against the whole training images to quantitatively evaluate the generation quantity.

\subsubsection{Implementation Details}\label{sec:exp_detail}
We employ StyleGAN2~\cite{karras2020analyzing_StyleGAN2} with adaptive discriminator augmentation~\cite{karras2020training_StyleGAN2-ADA} to build our ILSGAN. 
We simply adopt a generator with 4-channels output to generate foreground and mask layers at the same time: 3 channels for foreground and image and 1 channel for the mask. 
This 4-channels generator and background generator each contain a mapping network that maps 512-D latent code z to a 512-D intermediate latent code. 
Hyperparameters are consistently set across datasets: non-empty mask loss weight $\lambda_m=2$, mask size threshold $\eta=0.25$, mask binarization loss weight $\lambda_b=2$, and ILS-MI loss weight $\lambda_{\text{ils}}=1$.
Our ILSGAN is optimized using Adam optimizer with initial learning rate 0.0025 and beta parameters 0 and 0.99 until 8M images have been shown to the discriminator.
During training, we center crop the images to get square training inputs. In the testing phase, we select the checkpoint with lowest FID for evaluation. 
Following DRC~\cite{yu2021unsupervised_DRC}, we crop the images and masks according to ground truth object bounding boxes and resize them to square shape as test data.
Reported results in Sec.~\ref{sec:results} are conducted at 128$\times$128 resolution while all analytical experiments in Sec.~\ref{sec:exp_analysis} are conducted at 64$\times$64 resolution.
Experiments are conducted on single RTX 2080Ti GPU.

\begin{table*}[t]\centering
	\tablestyle{6pt}{1.2}\begin{tabular}{lrc cccc cccc cccc}
		\toprule
		\multirow{2}[2]{*}{~~Loss} & \multirow{2}[2]{*}{$\lambda_{\text{ils}}$} & \multirow{2}[2]{*}{~} & \multicolumn{4}{c}{Cars} & \multicolumn{4}{c}{CUB} & \multicolumn{4}{c}{Dogs} \\
		\cmidrule(lr){4-7} \cmidrule(lr){8-11} \cmidrule(lr){12-15}
		& & & ~FID$\downarrow$ & IoU$\uparrow$ & DICE$\uparrow$ & MI$\downarrow$
		& ~FID$\downarrow$ & IoU$\uparrow$ & DICE$\uparrow$ & MI$\downarrow$
		& ~FID$\downarrow$ & IoU$\uparrow$ & DICE$\uparrow$ & MI$\downarrow$ \\
		\cmidrule(lr){1-15}
		w/o ILS & $-$~ & ~ & 7.7 & 80.6 & 88.9 & 7.32 & 8.9 & 68.6 & 80.1 & 7.40 & 37.9 & 81.7 & 89.6 & 7.33\\
		\cmidrule(lr){1-15}
		ILS-L1  & 1.0 & ~ & 7.8 & 82.1 & 89.9 & 6.78 & \textbf{8.7} & 72.5 & 83.1 & 6.64 & 40.1 & 82.9 & 90.4 & 7.43\\
		\cmidrule(lr){1-15}
		\multirow{5}{*}{ILS-MI} 
		& 0.2 & ~ & 7.7 & 80.4 & 88.8 & 6.61 & 8.8 & 70.0 & 81.2 & 7.14 & 38.2 & 82.8 & 90.3 & 7.36\\
		& 0.5 & ~ & 8.0 & 82.2 & 89.9 & 6.98 & 8.8 & 71.3 & 82.2 & 7.03 & \textbf{37.4} & \textbf{84.0} & \textbf{91.1} & 6.80\\
		& * 1.0 & ~ & \textbf{7.3} & \textbf{83.5} & \textbf{90.8} & 6.09 & 8.9 & 72.3 & 82.9 & 6.70 & 38.7 & 83.8 & 90.9 & \textbf{6.75}\\
		& 2.0 & ~ & 11.5 & 78.5 & 87.6 & \textbf{6.06} & 9.0 & \textbf{73.7} & \textbf{84.0} & \textbf{6.35} & 40.8 & 80.3 & 88.7 & 7.05\\
		& 5.0 & ~ & 22.2 & 61.7 & 75.9 & 6.70 & 17.4 & 49.3 & 61.3 & 7.51 & 55.3 & 44.2 & 60.3 & 8.74\\
		
		\bottomrule
	\end{tabular}
	\caption{Quantitative evaluations of ILS losses ($\lambda_\text{ils}$ and * denotes the weight of loss and the default setting respectively).}
    \label{tab:weight_and_L1_ablation}
\end{table*}

\begin{table*}[t]\centering
	\tablestyle{4.3pt}{1.25}\begin{tabular}{rcccc cccc cccc cccc}
		\toprule
		
		 & \multirow{2}[2]{*}{Dist} & \multirow{2}[2]{*}{Sep?} & \multirow{2}[2]{*}{OptVis?} & \multirow{2}[2]{*}{OptM?} & \multicolumn{4}{c}{Cars} & \multicolumn{4}{c}{CUB} & \multicolumn{4}{c}{Dogs} \\
		\cmidrule(lr){6-9} \cmidrule(lr){10-13} \cmidrule(lr){14-17}
		& & & & & ~FID$\downarrow$ & IoU$\uparrow$ & DICE$\uparrow$ & MI$\downarrow$ 
		& ~FID$\downarrow$ & IoU$\uparrow$ & DICE$\uparrow$ & MI$\downarrow$
		& ~FID$\downarrow$ & IoU$\uparrow$ & DICE$\uparrow$ & MI$\downarrow$ \\
		\cmidrule(lr){1-17}
		A~~ & \multicolumn{4}{c}{w/o $\mathcal{L}_\text{ILS-MI}$} & 7.7 & 80.6 & 88.9 & 7.32 & 8.9 & 68.6 & 80.1 & 7.40 & 37.9 & 81.7 & 89.6 & 7.33 \\
		\cmidrule(lr){1-17}
		*B~~ & \multirow{4}{*}{$\mathcal{L}$}  & \checkmark & \checkmark & \checkmark & \textbf{7.3} & \textbf{83.5} & \textbf{90.8} & \textbf{6.09} & 8.9 & \textbf{72.3} & \textbf{82.9} & 6.70 & 38.7 & 83.8 & 90.9 & \textbf{6.75}\\
		C~~ & & \checkmark  & $\times$ & \checkmark & 7.5 & 79.6 & 87.8 & 6.88 & \textbf{8.6} & 70.9 & 82.0 & 7.05 & \textbf{37.6} & \textbf{84.4} & \textbf{91.3} & 7.22 \\
	    D~~ & ~  & \checkmark & \checkmark & $\times$ & 7.9 & 81.3 & 89.3 & 6.47 & 8.9 & 70.2 & 81.4 & \textbf{6.54} & 39.2 & 82.9 & 90.3 & 7.08\\
		E~~ & ~  & $\times$ & $-$ & $-$ & 8.1 & 82.0 & 89.7 & 6.45 & 9.1 & 69.5 & 80.9 & 6.92 & 39.0 & 82.9 & 90.4 & 7.04\\
		\cmidrule(lr){1-17}
		F~~ & $\mathcal{N}$  & \checkmark & \checkmark & \checkmark & 13.5 & 77.3 & 86.8 & 6.71 & 21.0 & 59.6 & 72.5 & 6.78 & 52.0 & 79.0 & 87.8 & 7.22\\
		\bottomrule
	\end{tabular}
	\caption{Ablation study of ILS-MI loss with respect to separate region (Sep?), optimizing visible region (OptVis?), and optimizing mask (OptM?). 
	$\mathcal{N}$ and $\mathcal{L}$ denote Gaussian and Laplace distribution, respectively. 
	* denotes the default setting.}
    \label{tab:ILS_ablation}
\end{table*}

\subsection{Benchmark Results}\label{sec:results}

Our method is compared with other related unsupervised and weakly-supervised foreground and background segmentation methods concerning segmentation performance and generation quality on Cars, CUB, and Dogs datasets. 
As GAN-based method is notoriously unstable in the training process, the performance of ILSGAN is reported as the average of three times experiments.
Table.~\ref{tab:compare_others} and Fig.~\ref{fig:generation_results} show the quantitative and qualitative results from which we have the following observations. 

(1) ILSGAN achieves superior unsupervised segmentation performance which surpasses competitors with large margins. On Cars and Dogs, ILSGAN outperforms previously the best method, DRC~\cite{yu2021unsupervised_DRC}, by 10.4/6.8 IoU/DICE and 14.0/8.9 IoU/DICE, respectively. On CUB, ILSGAN gains 2.4 IoU improvement over previously the best method, Impr.LGAN~\cite{yang2022learning_LFBS}.

(2) ILSGAN even significantly outperforms other layered GANs, OneGAN~\cite{benny2020onegan_OneGAN} and FineGAN~\cite{singh2019finegan_FineGAN}, which requires weak supervision such as bounding box annotation. Note that  ILSGAN respectively gains 10.0, 16.6, and 15.3 IoU improvement on Cars, CUB, and Dogs compared to OneGAN.  

(3) ILSGAN can synthesize images of high quality, achieving 9.0, 8.7, and 49.5 FID on Cars, CUB, and Dogs, respectively, either significantly outperforms or performs on par with other layered GANs such as FineGAN, OneGAN, and Impr.LGAN. Examples in Fig.~\ref{fig:generation_results} show the high fidelity and clear foreground-background disentanglement of the generated images.

\subsection{Analysis} \label{sec:exp_analysis}

\paragraph{The effect of ILS-MI} 
We study the effect of ILS-MI loss with its weight $\lambda_{\text{ils}}\in\{0.0, 0.2, 0.5, 1.0, 2.0, 5.0\}$, where $\lambda_{\text{ils}}=0.0$ corresponds to ``w/o ILS''.
The quantitative and qualitative results are presented in Table.~\ref{tab:weight_and_L1_ablation} and Fig.~\ref{fig:ablations_fig} respectively. 
Compared to the baseline method (w/o ILS), our method with ILS-MI ($\lambda_{\text{ils}}=1$) reduces MI by 1.2, 0.7, and 0.6 on Cars, CUB, and Dogs, respectively, showing improved layer independence. 
By visualization in Fig.~\ref{fig:ablations_fig}, while information leakage emerges in the results of the baseline method (\eg car wheels and texture are present in the background), the foreground and background layers in the results of our method have less visual confusion, suggesting information leakage issue is effectively suppressed by our ILS-MI.
Accordingly, our ILS-MI brings clear improvement in segmentation performance with 2.9 IoU on Cars, 3.7 IoU on CUB, and 2.1 IoU on Dogs. 
These results justify that pursuing independent layer synthesis with ILS-MI can suppress the information leakage and thereby significantly improve segmentation performance.

We also notice that raising $\lambda_{\text{ils}}$ too high has a risk of overwhelming the effect of other losses such as generation quality by adversarial loss and binary mask by binarization loss. 
It can be observed that the loss weight of ILS-MI greater than 2 leads to significantly degraded generation quality (high FID), non-binary foreground masks (Fig.~\ref{fig:ablations_fig}), and accordingly the segmentation performance decrease.
Therefore, we use a secure loss weight $\lambda_{\text{ils}}=1$ to pursue independent layers without toppling down the synthesis.
\paragraph{Comparison of ILS-MI and ILS-L1}

We further compare the effect of our ILS-MI and the intuitive alternative, ILS-L1.
The quantitative and qualitative results of ILS-L1 are also presented in Table.~\ref{tab:weight_and_L1_ablation} and Fig.~\ref{fig:ablations_fig}.
We observe that ILS-L1 can achieve similar effects to ILS-MI: it reduces MI (only on Cars and CUB), suppresses information leakage (Fig.~\ref{fig:ablations_fig}), and improves the segmentation performance. 
Despite so, ILS-L1 still shows the following drawbacks compared to ILS-MI.
(1) Fig.~\ref{fig:ablations_fig} shows that artifacts (\ie white block in the center of the background) emerge in the generated background layer of ILS-L1, possibly due to the invisible background being optimized towards the opposite color of visible foreground. In contrast, ILS-MI does not suffer from this problem.
(2) ILS-L1 generally brings less improvement in MI and segmentation performance compared to ILS-MI. 
These results show that ILS-MI, built upon directly minimizing MI upper bound, achieves more robust and better performance than the intuitive alternative ILS-L1. 

\begin{figure}[t]
   \centering
   \includegraphics[width=\linewidth]{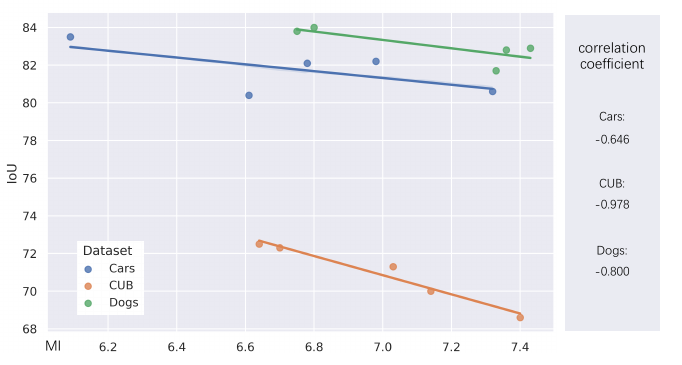}
   \caption{Correlation between IoU and MI. 
   The negative correlation suggests that reducing MI to increase independence contributes to segmentation. 
   }
   \label{fig:correlation_fig}
\end{figure}

\subsubsection{Correlation between MI and IoU}
We further plot the MI-IoU data collected from Table.~\ref{tab:weight_and_L1_ablation} and shows the correlation coefficients between MI and IoU in Fig.~\ref{fig:correlation_fig}.
The degraded results of which ILS-MI loss weight greater or equal to 2 are not included.
It is a clear to see negative correlation between MI and IoU across three datasets, justifying that layer independence can positively contribute to the segmentation performance.

\subsection{Ablation Study}
We conduct an ablation study about the concrete design choices in our ILS-MI losses. Results are summarized in Table.~\ref{tab:ILS_ablation} and visualizations are available in the appendix.

\paragraph{Visible region optimization}
We disable the optimization of visible regions by stopping gradients of ILS-MI loss with respect to $\rvf_{\text{vis}}$ and $\rvb_{\text{vis}}$ in Eq.~\ref{eq:ils_inv_vis}.
Results are shown in row C in Table.~\ref{tab:ILS_ablation}.
By comparing row B to row C, it can be seen that optimizing visible regions can decrease MI on all three datasets, has more robust segmentation performance, yet has no noticeable effect on synthetic fidelity.

\paragraph{Mask optimization}
We ablate the optimization of the mask by stopping the gradients of ILS-MI loss with respect to $\rvm$ in Eq.~\ref{eq:ils_inv_vis}. 
Results are shown in row D in Table.~\ref{tab:ILS_ablation}.
By comparing row B to row D, it is obvious that optimizing the mask can improve segmentation performance on all three datasets and reduce MI on Cars and Dogs, indicating that segmentation is critical to layer independence.

\paragraph{Separate regions}
We ablate the separation of invisible and visible regions in Eq.~\ref{eq:ils_inv_vis}, leading to a form of $\mathcal{L}_\text{ILS} = I(\rvb;\rvf)$ that computes the MI of the whole foreground and background layers.
As shown in row E of Table.~\ref{tab:ILS_ablation}, separate region, comparing no separation, achieves lower MI and has advantages of both synthetic quality and segmentation performance on all three datasets. This demonstrates the merits of our separation according to mask and visibility.

\paragraph{Distribution of $q(\rvb|\rvf)$}
Besides Laplace distribution, we also try Gaussian distribution to serve as the approximation of $q(\rvb|\rvf)$.
As shown in row F of Table.~\ref{tab:ILS_ablation}, Gaussian distribution completely fails when compared to Laplace and even worse than not using ILS. 

\section{Conclusion}
We propose Independent Layer Synthesis GAN, which generates independent foreground and background layers for fully unsupervised foreground-background segmentation. 
It improves interlayer independence to reduce the semantic and visual confusion caused by information leakage. 
To enhance independence, we minimize the MI between visible and invisible regions of foreground and background, resulting in more precise masks.
Experiments on various single object datasets demonstrate ILSGAN surpasses other unsupervised methods significantly on segmentation performance and synthetic quality. 
Moreover, the results show that ILSGAN performs more impressively and robustly compared to our intuitive approach, which only reduces visual similarity.
We hope our work will motivate future research in unsupervised segmentation, even self-supervised learning~\cite{he2022masked_MAE}, \textit{e.g.} foreground masks as attention.

\section*{Acknowledgments}
This work was supported by the National Key R\&D Program of China under Grant 2018AAA0102801.

\bibliography{aaai23}

\clearpage

\appendix

\section{Evaluation of MI}\label{sec:eval_mi}

\subsection{MINE Estimator}

When evaluating ILSGAN, we employ a heavier neural estimator, MINE~\cite{belghazi2018mine_MINE}, to more accurately measure the independence between synthesized foreground and background layers.
\begin{equation}\label{equ:mine_estimator}
\begin{aligned}
       \mimine(\rvb,\rvf, \theta) = & \frac{1}{N}\sum_{i=1}^{N} \left[ T_{\theta}(\rvb^{(i)},\rvf^{(i)}) \right] \\
      - &\log \left( \frac{1}{N}\sum_{i=1}^{N} e^{T_{\theta}(\rvb^{(i)},\rvf^{(k_i)})} \right),
\end{aligned}
\end{equation}
where $T_{\theta}$ is a learnable function implemented with an MLP. 
MINE is a lower bound of mutual information.
By optimizing $T_{\theta}$ to maximize Equ.~\ref{equ:mine_estimator}, the lower bound can serve as a reasonable estimated MI.
Similar to the ILS loss, we respectively compute the MI of regions divided by segmentation masks and then sum them up,
\begin{equation}
    \widehat{\text{MI}} = \mimine(\rvbginvis, \rvfgvis, \theta_1) + \mimine(\rvbgvis, \rvfginvis, \theta_2 ),
\end{equation}
where $\theta_1$ and $\theta_2$ are the parameters of two estimators respectively, and $\widehat{\text{MI}}$ is reported as "MI" in our work.

\subsection{Implementation Details}

We implement $T_{\theta}$ with an two-layer MLP. It directly takes two 64$\times$64$\times$3 layers (foreground and background layers) as input and the hidden dimension is set to 3000.
%
When optimizing the estimators, we set the batch size to 64 and use Adam optimizer with an initial learning rate of 0.0002 and beta parameters of 0.9 and 0.999. 
We use the ``OnPlateau" scheduler to adjust the learning rate for better convergence. The estimator is trained for 100K iterations. For every 400 iterations, we estimate the MI on our synthetic samples. And we average the results of the last 20 estimations as the reported MI.

\section{Training Details}\label{sec:training_details}

\paragraph{ILSGAN}

During the training of ILSGAN, we set the batch size to 32. And in order to stabilize the adversarial learning, we set the weight of ILS loss to 0.2 at the initial stage (0 to 400K).
%
We also use adaptive discriminator augmentation (ADA)~\cite{karras2020training_StyleGAN2-ADA} to facilitate the training. The target of ADA is set to 0.6, and we use augmentation of pixel blitting, geometric transformation, and color transformations in ADA.
We disable path length regularization~\cite{karras2019style_StyleGAN1} to stabilize the layered training.
%
For ILS-MI loss, we normalize it according to the pixel number, which can be simply implemented by (f-b).abs().mean() in PyTorch code when computing the $L1$ norm. And for the foreground and background layers' pixels of which the value is out of [-1,1], ILS loss does not optimize them.

\paragraph{Segmentation}

To evaluate the segmentation performance of our methods, we train a U-Net~\cite{ronneberger2015u_U-Net} on 10K synthetic samples from ILSGAN.
The U-Net is trained for 6000 iterations with 64 batch size and is optimized with Adam optimizer. We use ``StepLR" learning rate scheduler, which has a learning rate of 0.001 during 0 to 4000 iterations and a 0.0002 learning rate during 4000 to 6000 iterations.
We apply color augmentation to input samples during training, which randomly does luma flip and adjusts the brightness, contrast, hue, and saturation.
We split 10K synthetic samples into 9K for training and 1K for validating. The U-Net is evaluated on validation set for every 1000 iterations. The best-performing checkpoint on validation set will be evaluated on the test set to get the segmentation performance.

\section{Additional Details of ILS Loss}\label{sec:ILS_loss}

\subsection{ILS-L1 Loss}

In the main text, we propose the ILS-L1 loss as follows,
\begin{equation}
    \lossILSLOne = - \| \rvbginvis - \rvfgvis \|_1 
     - \| \rvbgvis - \rvfginvis \|_1.
\end{equation}
We compare the settings with or without gradients back to the visible regions of foreground and background to investigate the optimization of visible regions.
\begin{figure*}[t]
    \centering
    \includegraphics[width=\linewidth]{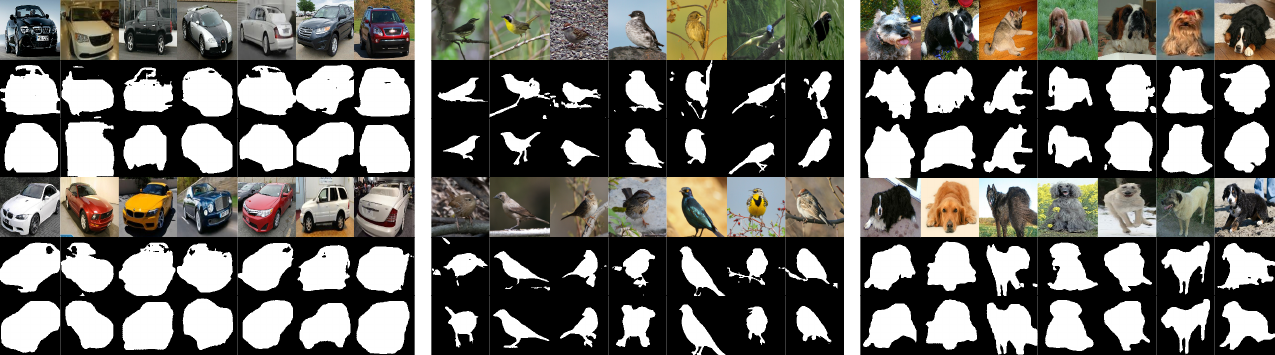}
    \caption{Qualitative segmentation results (128$\times$128) on Cars (left), CUB (middle), and Dogs (right). From top to bottom of each sample: real image, predicted mask, and ground-truth mask. Note that the ground-truth of Cars and Dogs are approximated ground-truth by DRC~\cite{yu2021unsupervised_DRC}.}
    \label{fig:appendix_viz_seg128}
\end{figure*}
\begin{table*}[t]\centering
	\tablestyle{6pt}{1.5}\begin{tabular}{l cccc cccc cccc}
		\toprule
		\multirow{2}[2]{*}{~~Loss} & \multicolumn{4}{c}{Cars} & \multicolumn{4}{c}{CUB} & \multicolumn{4}{c}{Dogs} \\
		\cmidrule(lr){2-5} \cmidrule(lr){6-9} \cmidrule(lr){10-13}
		& ~FID$\downarrow$ & IoU$\uparrow$ & DICE$\uparrow$ & MI$\downarrow$
		& ~FID$\downarrow$ & IoU$\uparrow$ & DICE$\uparrow$ & MI$\downarrow$
		& ~FID$\downarrow$ & IoU$\uparrow$ & DICE$\uparrow$ & MI$\downarrow$ \\
		\cmidrule(lr){1-13}
		$\lossILSLOneSingle$ & \textbf{7.4} & \textbf{82.1} & 89.8 & 7.16 & 9.0 & 71.1 & 82.1 & 7.02 & \textbf{39.0} & 80.3 & 88.6 & \textbf{6.99}\\
		$\lossILSLOneSeparate$ & 7.8 & \textbf{82.1} & \textbf{89.9} & \textbf{6.78} & \textbf{8.7} & \textbf{72.5} & \textbf{83.1} & \textbf{6.64} & 40.1 & \textbf{82.9} & \textbf{90.4} & 7.43\\
		\bottomrule
	\end{tabular}
	\caption{Quantitative comparison of $\lossILSLOneSingle$ and $\lossILSLOneSeparate$.}
    \label{tab:ILS_L1_ablation}
\end{table*}
With the gradients back to visible regions, ILS-L1 loss is equal to maximize the $L1$ distance between whole foreground and background layers,
\begin{equation}
\begin{aligned}
    \lossILSLOneSingle &= - \| \rvbginvis - \rvfgvis \|_1 
     - \| \rvbgvis - \rvfginvis \|_1. \\
     &= - \| \rvb \odot \rvmask - \rvf \odot \rvmask \|_1 \\
     &~~~~~~~~~~~~~~~~ - \| \rvb \odot (\bm{1} - \rvmask) - \rvf \odot (\bm{1} - \rvmask) \|_1. \\
     &= - \| (\rvb - \rvf) \odot \rvmask \|_1 - \| ( \rvb - \rvf ) \odot (\bm{1} - \rvmask) \|_1. \\
     &= - \| \rvb - \rvf \|_1. \\
\end{aligned}
\end{equation}
When disabling the gradients back to visible regions of foreground and background, it can construct separate regions similar to ILS-MI,
\begin{equation}
\begin{aligned}
    \lossILSLOneSeparate &= - \| \rvb \odot \rvmask - \overline{\rvf} \odot \rvmask \|_1 \\
     &~~~~~~~~~~~~~~~~ - \| \overline{\rvb} \odot (\bm{1} - \rvmask) - \rvf \odot (\bm{1} - \rvmask) \|_1. \\
     &= - \| (\rvb - \overline{\rvf}) \odot \rvmask \|_1 - \| ( \overline{\rvb} - \rvf ) \odot (\bm{1} - \rvmask) \|_1.
\end{aligned}
\end{equation}
In both settings, the gradients back to the mask are accumulated to zero, meaning that the mask layer will not be optimized by ILS-L1 loss.

The quantitative results of $\lossILSLOneSingle$ and $\lossILSLOneSeparate$ are reported in Table.~\ref{tab:ILS_L1_ablation}.
We report the results of $\lossILSLOneSeparate$ in the main text.

\subsection{Approximations of \texorpdfstring{$q(\rvb|\rvf)$}{TEXT} in ILS-MI loss}

ILS-MI loss minimizes the CLUB upper bound to indirectly reduce MI. Given a mini-batch of samples $\{(\rvb^{(i)},\rvf^{(i)})\}_{i=1}^N$, its unbiased $N$-sample form writes
\begin{equation}
    \miclub(\rvb,\rvf) = \frac{1}{N}\sum_{i=1}^{N} \log{ \frac{q(\rvb^{(i)}|\rvf^{(i)})}{q(\rvb^{(k_i)}|\rvf^{(i)})} },
\end{equation}
where the $q(\rvb|\rvf)$ needs to be approximated.

\paragraph{Non-parametric approximation}
In the main text, we propose to approximate $q(\rvb|\rvf)$ with Laplace distribution $q(\rvb|\rvf) = \mathcal{L}(\rvb;\rvf, \mathbf{I}) \propto \exp(-\| \rvb - \rvf \|_1)$,
\begin{equation}
    \miclub(\rvb,\rvf) = \frac{1}{N}\sum_{i=1}^{N} - \| \rvb^{(i)} - \rvf^{(i)} \|_1 + \| \rvb^{(k_i)} - \rvf^{(i)} \|_1,
\end{equation}
And the approximation with Gaussian distribution $\mathcal{N}(\rvb|\mu=\rvf,\sigma=\mathbf{I})$ reported in the ablation study is derived as follows,
\begin{equation}
    \miclub(\rvb,\rvf) = \frac{1}{N}\sum_{i=1}^{N} \frac{1}{2} \left\{ - ( \rvb^{(i)} - \rvf^{(i)} )^2 + ( \rvb^{(k_i)} - \rvf^{(i)} )^2 \right\}
\end{equation}
According to the experiment results in the main text, ILS-MI loss with Laplace distribution achieves a more robust and better performance.

\paragraph{Approximation with neural network}
\begin{figure*}[t]
    \centering
    \includegraphics[width=\linewidth]{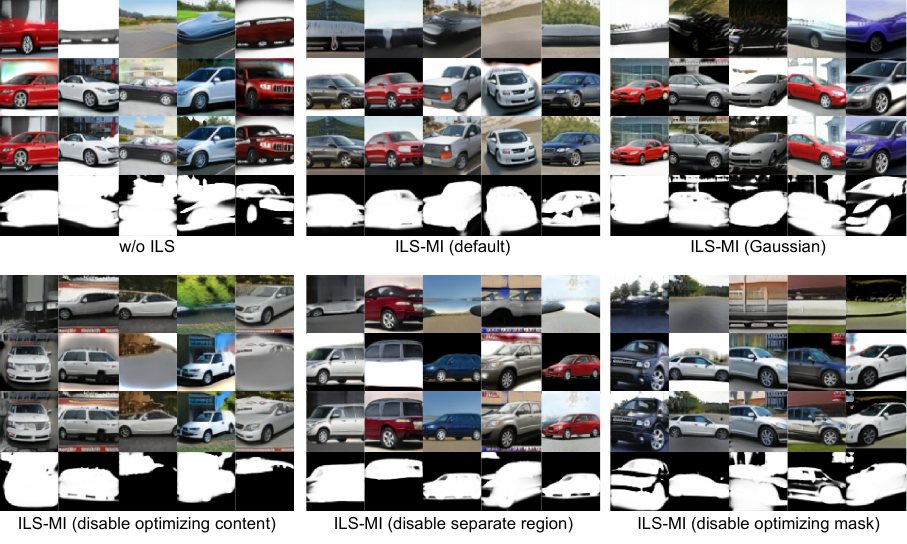}
    \caption{Qualitative results for ablation study of ILS loss. From top to bottom of each sample: background, foreground, composed image, and corresponding mask.}
    \label{fig:appendix_ablation}
\end{figure*}
We report the additional results of estimating $q(\rvb|\rvf)$ with neural networks to learn the parameters of Laplace and Gaussian distribution respectively.
When approximated with Laplace distribution, we use neural networks $\mu_\theta$ and $b_\theta$ to learn $\mu$ and $b$, $q(\rvb|\rvf) = \mathcal{L}(\rvb;\mu=\mu_\theta(\rvf), b=b_\theta(\rvf)) \propto \exp(- \left\| \frac{ \rvb - \mu_\theta(\rvf) }{b_\theta(\rvf)} \right\|_1 )$. With the approximation, our ILS loss is as following,
\begin{equation}
\begin{aligned}
    \miclub(\rvb,\rvf) = \frac{1}{N}\sum_{i=1}^{N} &- \left\| \frac{ \rvb^{(i)} - \mu_\theta(\rvf^{(i)}) }{b_\theta(\rvf^{(i)})} \right\|_1 \\
    &~~~~~~~~ + \left\| \frac{ \rvb^{(k_i)} - \mu_\theta(\rvf^{(i)}) }{b_\theta(\rvf^{(i)})} \right\|_1.
\end{aligned}
\end{equation}
We minimize the negative log-likelihood to learn $\mu_\theta$ and $b_\theta$,
\begin{equation}\label{equ:NLL_Laplace}
    \mathcal{L}_{NLL} = \left\| \frac{ \rvb - \mu_\theta(\rvf) }{b_\theta(\rvf)} \right\|_1,
\end{equation}
where the constant term is omitted in Equ.~\ref{equ:NLL_Laplace}.
Similar to learning the Laplace distribution, when approximated with Gaussian distribution, we also use neural network $\mu_\theta$ and $\sigma_\theta$ to learn $\mu$ and $\sigma$ in $\mathcal{N}(\rvb|\mu=\mu_\theta(\rvf),\sigma=\sigma_\theta(\rvf) \odot \mathbf{I})$ and optimize the parameters by minimizing NLL.
%
%
However, during adversarial learning, when plugging in the networks which learn the distribution, it fails the learning of mask layer and even collapses the adversarial learning. This failure may be caused by the instability from introducing one more network in the training of GAN. Thus our ILSGAN adopts the approximation with non-parametric Laplace distribution.

\section{Qualitative Results}\label{sec:qualitative_results}

We provide the qualitative results in Fig.~\ref{fig:appendix_ablation} corresponding to the ablation study in the main text,
and visualization of evaluated segmentation results in Fig.~\ref{fig:appendix_viz_seg128} corresponding to the 128$\times$128 experiments in the main text.

\end{document}